\documentclass{article} 
\usepackage{iclr2026_conference,times}

\usepackage{amsmath,amsfonts,bm}

\def\eqref#1{equation~\ref{#1}}

\def\1{\bm{1}}

\DeclareMathAlphabet{\mathsfit}{\encodingdefault}{\sfdefault}{m}{sl}
\SetMathAlphabet{\mathsfit}{bold}{\encodingdefault}{\sfdefault}{bx}{n}

\usepackage{hyperref}
\usepackage{url}
\usepackage{graphicx}
\usepackage{booktabs}
\usepackage{wrapfig}
\usepackage{array}
\usepackage{multirow}
\usepackage{makecell}
\usepackage{enumitem}
\usepackage{amsmath,amssymb}
\usepackage[nameinlink]{cleveref}
\usepackage{float}
\newcommand{\will}[1]{}

\title{Are You Getting What You Pay For? Auditing Model Substitution in LLM APIs}

\author{Will Cai,~~Tianneng Shi,~~Xuandong Zhao\thanks{Corresponding author},~~Dawn Song \\
University of California, Berkeley\\
\texttt{\{wicai,stneng,xuandongzhao,dawnsong\}@berkeley.edu} 
}

\iclrfinalcopy
\begin{document}

\maketitle

\begin{abstract}
Commercial Large Language Model (LLM) APIs create a fundamental trust problem: users pay for specific models but have no guarantee that providers deliver them faithfully. Providers may covertly substitute cheaper alternatives (e.g., quantized versions, smaller models) to reduce costs while maintaining advertised pricing. We formalize this \emph{model substitution} problem and systematically evaluate detection methods under realistic adversarial conditions. Our empirical analysis reveals that software-only methods are fundamentally unreliable: statistical tests on text outputs are query-intensive and fail against subtle substitutions, while methods using log probabilities are defeated by inherent inference nondeterminism in production environments. We argue that this verification gap can be more effectively closed with hardware-level security. We propose and evaluate the use of Trusted Execution Environments (TEEs) as one practical and robust solution. Our findings demonstrate that TEEs can provide provable cryptographic guarantees of model integrity with only a modest performance overhead, offering a clear and actionable path to ensure users get what they pay for. Code is available at \url{https://github.com/sunblaze-ucb/llm-api-audit}
\end{abstract}

\section{Introduction}

Large Language Models (LLMs) have demonstrated remarkable capabilities, leading to widespread adoption through cloud-based APIs \citep{openai_chatgpt, anthropic_claude, google_gemini, together_inference}. Enterprises and researchers select and pay for specific models (e.g., Llama 405B vs 70B) based on advertised capabilities, performance on leaderboards, and expected behavior \citep{touvron2023llama, qwen, mistral2023mistral7b}. However, this black-box access model operates on an implicit assumption of trust: that the provider will faithfully serve the requested model. The immense computational cost of hosting state-of-the-art LLMs creates a powerful economic incentive for providers to violate this trust through \emph{model substitution}—covertly replacing the advertised model with a cheaper, less powerful alternative.

This substitution threat is not hypothetical. A provider might swap a flagship model for a smaller variant or a heavily quantized version (e.g., INT8/FP8) to reduce GPU costs and increase profit margins \citep{modelequalitytesting, sun2024svip}. Substitutions can also occur for operational reasons, such as rerouting traffic from an overloaded high-end model to a less-utilized, lower-tier one (e.g., a version fine-tuned on different data). While the intent may not always be malicious, any undisclosed substitution breaks the service agreement, compromises the reliability of dependent applications, hinders reproducible research, and invalidates the benchmark results. Furthermore, subtle changes from quantization or fine-tuning might inadvertently affect model safety or introduce biases \citep{hong2024decoding}. Imagine a researcher relying on a specific model's advertised capabilities for scientific analysis, only to unknowingly receive results from a substituted, less capable model, potentially invalidating their findings.

For a user or auditor, verifying the model behind a black-box API is fundamentally challenging. The interaction is typically limited to sending prompts and receiving text (and optionally, token log probabilities~\citep{stealingproductionlanguagemodel}), with no direct access to the model's weights or the underlying infrastructure. This \emph{information asymmetry} heavily favors the provider. A sophisticated provider can exploit this by employing active countermeasures, such as detecting and routing benchmark queries to the genuine model while serving a substitute to regular traffic, or by randomly mixing outputs from different models to obscure the statistical signal of the substitution.

This paper confronts the critical problem of auditing model substitution in black-box LLM APIs. We formalize the verification task, analyze the limitations of existing software-based methods under realistic adversarial conditions, and present a robust solution. Our central thesis is that the inherent ambiguity and nondeterminism of software-level signals make them insufficient for reliable auditing. Instead, we argue that hardware-backed attestation via Trusted Execution Environments (TEEs) is the only currently viable mechanism to achieve strong, efficient, and provable integrity for LLM APIs. Our contributions are:
\begin{itemize}[leftmargin=*, ]
    \item We formalize the problem of LLM model substitution detection from the perspective of a black-box user/auditor.
    \item We design and analyze practical adversarial scenarios for model substitution, including quantization, randomized substitution, and benchmark evasion.
    \item We empirically evaluate the effectiveness of existing detection techniques, showing that text-based statistical tests lack power against subtle or randomized substitutions, while metadata-based methods like log probability comparison are defeated by production-level inference nondeterminism.
    \item We propose and evaluate TEEs as a practical and deployable solution, demonstrating that they uniquely provide provable model integrity in black-box APIs. We show that TEEs offer strong security guarantees with only a modest performance cost, establishing them as the most actionable path toward substitution-resistant LLM services.
\end{itemize}

\section{Related work}
\label{sec:related_work}

\textbf{LLM API monitoring and auditing. }
Several studies have monitored the behavior and performance of commercial LLM APIs over time. \citet{chen2023chatgptsbehaviorchangingtime} track changes in ChatGPT's capabilities, highlighting behavioral drift, while \citet{eyuboglu2024model} characterizes updates to API-accessed ML models. Closer to our work, \citet{modelequalitytesting} specifically test API output text distributions against reference distributions using Maximum Mean Discrepancy (MMD). Other approaches include direct auditing via identity-style prompting \citep{mitigatingadversarialmanipulation} and predicting black-box LLM performance using self-queries \citep{sam2025predicting}. When model internals are available, TopLoc \citep{ong2025toploclocalitysensitivehashing} uses locality-sensitive hashing over intermediate activations to produce proofs of correct execution. While these methods provide a foundation for auditing, their robustness against determined adversaries and in realistic, non-deterministic production environments \citep{he2025nondeterminism} remains an open question, which we address in this work.

\textbf{Detecting LLM-generated text. } Detection of LLM-generated text has been extensively studied, including post-hoc methods and proactive watermarking techniques \citep{yang2023survey, ghosal2023towards}. Zero-shot detection approaches use stylistic features or model-specific "fingerprints" to distinguish AI-generated content without specialized training data \citep{mitchell2023detectgpt, bao2023fast, yang2024dna}. Trained classifiers \citep{hu2023radar, idiosyncrasieslargelanguagemodels} utilize datasets from various sources to differentiate model outputs. Although effective against stylistically distinct models, these approaches may struggle with subtle substitutions within the same model family. LLM watermarking techniques embed hidden signals in outputs to trace content ownership \citep{kirchenbauer2023watermark, zhao2024provable, zhao2024sok}, but they are provider-centric and not intended for end-user verification of model \emph{identity}, particularly as the same watermark might span different backend models.

\textbf{Verifiable computation for ML. } Cryptographic and hardware-based techniques can provably verify ML inference. Zero-Knowledge Proofs (ZKPs) allow proof of correct inference without revealing inputs or model weights but face substantial computational costs for large LLMs, making them impractical for real-time APIs today \citep{sun2024zkllm, xie2025zkpytorch}. In contrast, Trusted Execution Environments (TEEs) provide hardware-level guarantees of integrity and confidentiality with much lower overhead \citep{nvidia}. Our work builds on this insight, arguing that TEEs represent the most mature and practical path toward verifiable LLM inference in the \emph{current} ecosystem.

\section{Problem formulation and threat models}
\subsection{Problem formulation}
\label{sec:problem_formulation}

Consider an LLM API service with three entities:
\begin{itemize}[leftmargin=*,nosep]
    \item \textbf{User/Auditor:} Use LLM services for tasks with input prompts \(x\) drawn from a distribution \(\pi(x)\).
    \item \textbf{Service provider:} Offers access to an LLM via a black-box API. The provider claims to serve a specific target model $M_{\text{spec}}$ but may substitute it with $M_{\text{alt}}$.
    \item \textbf{Target model (\(M_\text{spec}\)):}  The advertised model with distribution $P_{\text{spec}}(y|x)$.
\end{itemize}

\textbf{Model substitution } occurs when the provider's actual backend distribution $P_{\text{actual}}(y|x)$ differs from the advertised $P_{\text{spec}}(y|x)$ \emph{without user disclosure}. Substitutions include: (1) smaller models (e.g., 7B vs 70B), (2) quantized variants (INT8 vs FP16) of $M_\text{spec}$, (3) fine-tuned or updated versions with altered training data or objectives, or (4) entirely different model families.

\textbf{Open-source vs. proprietary access. } 
The user's ability to audit depends on the type of model and API. For \emph{open-source models}, an auditor can run \(M_\text{spec}\) locally to get reference outputs and log probabilities. For \emph{proprietary models} (e.g., GPT-4, Claude-4), the auditor only has black-box access, creating a significant information asymmetry.

\textbf{Auditing goal.} The user/auditor aims to determine whether the provider is faithfully using the specified model. Formally, given samples \((x, y)\) from the provider's API (\(y \sim P_\text{actual}(\cdot|x)\)) and potentially reference samples from \(P_\text{spec}(\cdot|x)\), the auditor wants to test the null hypothesis:
\[
H_0: P_\text{actual}(y|x) = P_\text{spec}(y|x) \quad \forall x \sim \pi(x) \quad \text{(Honest Provider)}
\]
against the alternative hypothesis:
\[
H_1: P_\text{actual}(y|x) \neq P_\text{spec}(y|x) \quad \text{for some } x \sim \pi(x) \quad \text{(Substitution Occurred).}
\]
An effective verification method should achieve: \emph{(1) Reliability} - low false positive/negative rates, \emph{(2) Efficiency} - reasonable query budgets, \emph{(3) Robustness} - resistance to adversarial countermeasures, and \emph{(4) Generality} - applicability across model types and API configurations.

\subsection{Adversarial attack scenarios}
\label{sec:attacks}

\textbf{Quantization substitution. } The provider replaces the full-precision target model \(M_\text{spec}\) with a quantized version (e.g., INT8, FP8, NF4). Quantization significantly reduces memory footprint and often accelerates inference, lowering costs. While preserving much of the model's capabilities, it slightly alters the output distribution \(P_\text{alt}(y|x)\). The attack relies on this distributional difference being too small for naive detection methods to pick up, especially with limited samples.

\textbf{Randomized model substitution. }
To make detection harder, the provider can route a query to a cheaper substitute \(M_\text{alt}\) with probability \(p\) and to the specified model \(M_\text{spec}\) with probability \(1-p\). The observed distribution is
$
P_\text{mixed}(y \mid x) = p \cdot P_\text{alt}(y \mid x) + (1-p) \cdot P_\text{spec}(y \mid x).
$
As \(p\to1\) (low substitution rate), \(P_\text{mixed}\) approaches \(P_\text{spec}\), making sampling-based detection increasingly difficult. A sophisticated provider might adapt \(p\) downward only when traffic appears non-audit-like.

\textbf{Benchmark evasion (cached/routed output). } This attack targets verification methods relying on known, fixed prompts, such as benchmark datasets or identity queries. The provider detects likely audit queries (e.g., via hashing or embedding similarity) and routes them to a genuine \(M_\text{spec}\) instance or returns cached outputs and metadata. Ordinary traffic is served by \(M_\text{alt}\).

\textbf{Limiting information disclosure. }
Providers might proactively limit the information exposed via APIs (e.g., removing full logits, restricting top-$k$ log probabilities) after demonstrations that such features can leak sensitive details \citep{stealingproductionlanguagemodel, protectedllmsleak}.

\section{Model verification techniques and robustness analysis}
\label{sec:verification_methods}

\begin{table}[t]
    \centering
    \vspace{-10pt}
    \renewcommand{\arraystretch}{1.2}
    \resizebox{0.68\textwidth}{!}{
    \begin{tabular}{lccc}
    \toprule
        \textbf{Service Provider} & \textbf{Open Source Models} & \textbf{Decoding Parameters} & \textbf{Logprobs Output} \\
        \midrule
        Anyscale & Yes & Full Control & Yes \\
        Together.ai & Yes & Full Control & Yes \\
        Hugging Face & Yes & Full Control & Top 5 \\
        AWS Bedrock & Yes & Full Control & Yes \\
        Nebius AI & Yes & Full Control & Yes \\ 
        Vertex AI & Yes & Full Control & Top 5 \\
        Mistral & Yes & Partial Control & No \\
        DeepSeek & Yes & Full Control & Top 20 \\
        OpenAI & No & Partial Control & Depends on model \\
        Cohere & No & Full Control & Top 1 \\
        Anthropic & No & Full Control & No \\
        \bottomrule
    \end{tabular}
    }
    \vspace{-5pt}
    \caption{Transparency and control across LLM API providers (September 2025). ``Full Control'' implies typical parameters like temperature, top-p, top-k, etc. Log probability availability and limits vary by model/version.}
    \label{tab:service_providers}
    \vspace{-10pt}
\end{table}


LLM service providers offer a wide range of interfaces, from basic text-only web chats to advanced APIs that allow control over decoding parameters and access to token log probabilities (\Cref{tab:service_providers} and \Cref{tab:service_providers_doc}). The level of information disclosure significantly constrains which verification techniques can be applied. In this section, we systematically evaluate various auditing methods across different access levels: (1) text-only output, (2) output plus metadata (e.g., log probabilities, activations), and (3) provider-supplied integrity (e.g., TEEs in \Cref{sec:tee}). For each technique, we detail the methodology, describe our experimental setup for evaluating its robustness against the adversarial scenarios outlined in \Cref{sec:attacks}, and analyze its effectiveness.

\subsection{Text-output-based verification}
\label{sec:text_methods}

\subsubsection{Text classifier}

\textbf{Method. }
\citet{idiosyncrasieslargelanguagemodels} use a trained classifier to predict the source model of generated samples, exploiting stylistic “fingerprints” unique to each model. Given a dataset of model completion labeled with model, the classifier is trained to predict the label using cross-entropy loss.

\begin{wraptable}{r}{0.65\textwidth}
    \centering
    \vspace{-10pt}
    \renewcommand{\arraystretch}{1.2}
    \resizebox{0.60\textwidth}{!}{
    \begin{tabular}{lcccc}
    \toprule
         \textbf{Model} & \textbf{BERT Acc} & \textbf{T5 Acc} & \textbf{GPT2 Acc} & \textbf{LLM2Vec Acc}\\ \hline
         Llama3-70B-Instruct-FP8 & 50.55 & 50.10 & 50.45 & 49.90 \\
         Llama3-70B-Instruct-INT8 & 51.60 & 49.90 & 51.30 & 50.25 \\  \hline
         Gemma2-9b-it-FP8 & 49.95 & 50.30 & 51.20 & 49.80 \\
         Gemma2-9b-it-INT8 & 49.00 & 49.70 & 51.65 & 49.55 \\ \hline
         Mistral-7b-v3-Instruct-FP8 & 50.55 & 49.75 & 48.70 & 48.75 \\
         Mistral-7b-v3-Instruct-INT8 & 49.70 & 50.75 & 50.50 & 51.15 \\ \hline
         Qwen2-72B-Instruct-FP8 & 50.05 & 50.25 & 48.75 & 49.80 \\
         Qwen2-72B-Instruct-INT8 & 50.75 & 50.55 & 49.45 & 50.20 \\
         \bottomrule
    \end{tabular}
    }
    \caption{Binary classification accuracy between outputs of quantized and original models on UltraChat \citep{ding2023enhancing}. Accuracies near 50\% indicate failure to differentiate.}
    \label{tab:classifier_results}
    \vspace{-10pt}
\end{wraptable}

\textbf{Attack: Quantization substitution. }
We test whether classifiers could distinguish full-precision from quantized variants of Llama-3.1-70B, Gemma-2-9B, Mistral-7B-v0.3, and Qwen2-72B. Using UltraChat prompts (temperature $0.6$), we train on 10k samples per variant (with 1k validation) and evaluate classifiers with BERT, T5, GPT-2, and LLM2Vec embeddings.

\textbf{Analysis. }
Classification accuracies stay near chance across all model pairs and embedding backbones (Table~\ref{tab:classifier_results}), leaving us unable to reject $H_0$ that $M_{\text{spec}}$ and $M_{\text{alt}}$ produce indistinguishable outputs; deviations from 50\% can be attributed to sampling noise. The result is consistent across families, parameter sizes, and INT8/FP8 variants, and persists under alternative train/validation splits. In contrast to \citet{idiosyncrasieslargelanguagemodels}, who report around 98\% cross-model and 80\% cross-model size accuracy, the signal here is too weak to separate quantized from full-precision variants. We therefore find that text-based classifiers, even with strong encoders (BERT/T5/GPT-2/LLM2Vec), are ineffective for detecting quantization-based substitution under realistic sampling budgets.

\subsubsection{Identity prompting}

\textbf{Method.}
Prompt the model to disclose its identity (e.g., “Who are you?”, “Who developed you?”) and string-match against known identifiers (family, developer), following the simple classifier setup in \citet{mitigatingadversarialmanipulation}.

\textbf{Attack: Quantization substitution. }
We test whether quantization changes self-identification: for Llama3.1-70B-Instruct, Gemma2-9b-it, Mistral-7b-v3-Instruct, and Qwen2-72B-Instruct, we issue 500 identity queries per model/variant at temperature 0.6 and matched family (e.g., “Llama”) and developer (e.g., “Meta”).

\textbf{Attack: System prompt override. }
In this setting, we explicitly set the system prompt to mislabel the model (e.g., “\emph{You are LLaMA, an AI assistant developed by Meta AI}”) and then re-ran identity queries on Gemma2-9b-it, Mistral-7b-v3-Instruct, and Qwen2-72B-Instruct.

\textbf{Analysis. }
The results for quantization substitution remain consistent: models typically disclose only broad identifiers such as the family or developer, and rarely reveal specific details like parameter count or quantization status. Queries asking for direct specifications are successful in less than 5\% of cases, often resulting in refusals or incorrect responses from the model. When the system prompt is overridden, most models fail to correctly identify themselves, with accuracy dropping to nearly 0\%. Instead, they often adopt the identity given by the system prompt, although some contradictions occasionally occur (e.g., restating the true developer). Overall, these findings indicate that identity prompting is ineffective for detecting substitutions within the same family and is highly sensitive to changes in the prompt, making it an unreliable method for detecting substitutions in practice.

\subsubsection{Model equality testing}

\textbf{Method. }
When only text outputs are available, one can test whether two models induce the same distribution over completions by applying a Maximum Mean Discrepancy (MMD) test \citep{modelequalitytesting}. Let \(P\) and \(Q\) denote the distributions of two candidate models,
\[
\mathrm{MMD}^2(P, Q) \;=\; 
\mathbb{E}_{x,x' \sim P}[k(x,x')] \;+\; \mathbb{E}_{x,x' \sim Q}[k(x,x')] \;-\; 2 \mathbb{E}_{x \sim P, x' \sim Q}[k(x,x')],
\]
with the Hamming kernel
\(
k_{\mathrm{hamming}}(x,x') = \sum_{i=1}^L \mathbf{1}\{x_i = x'_i\}.
\)
We estimate significance by permutation testing: repeatedly shuffle completions between groups, recompute MMD, and reject \(H_0\) if the observed statistic exceeds the \(\alpha\)-quantile of this null distribution.

\begin{figure}[t]
    \centering
    \includegraphics[width=0.85\linewidth]{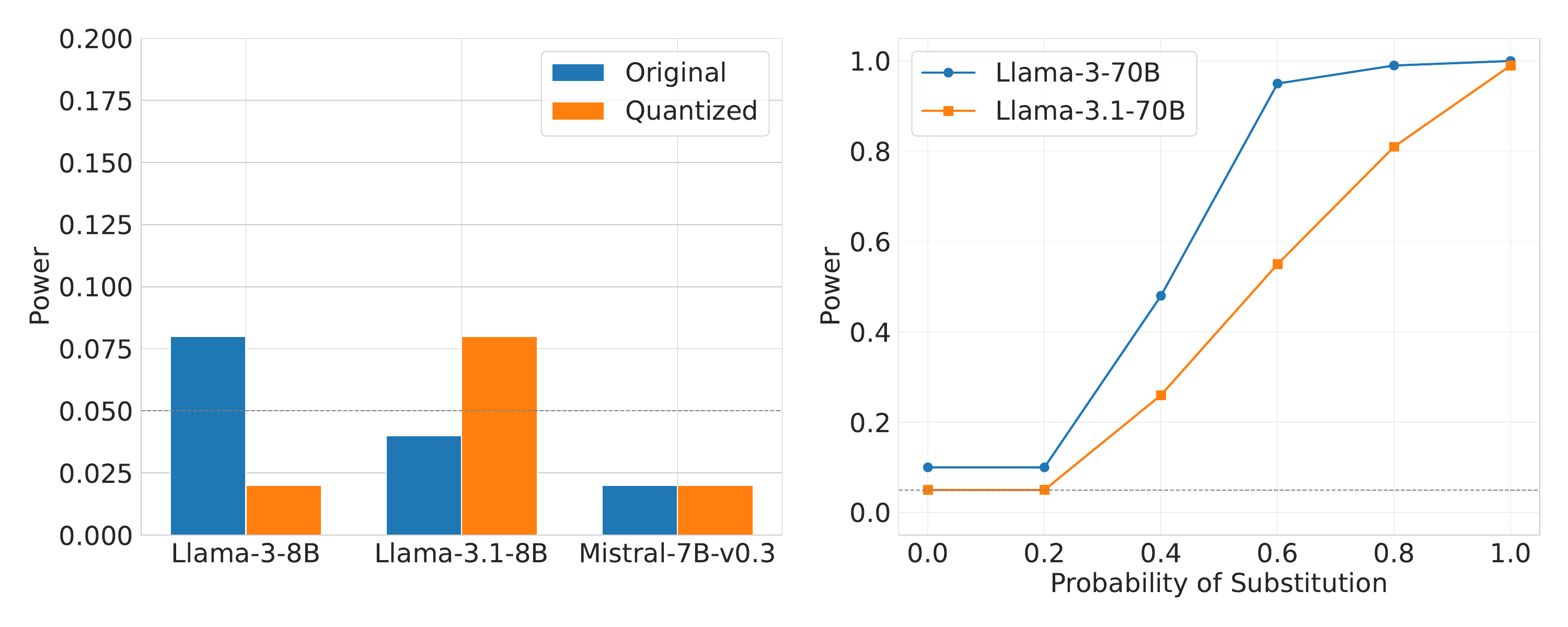}
    \vspace{-8pt}
    \caption{Statistical power of MMD tests. 
    (Left) Distinguishing quantized from original models when parameters are small. 
    (Right) Power as a function of substitution probability \(p\) for larger models, showing how mixtures of quantized and full-precision outputs affect detectability.}
    \label{fig:Model_equality}
    \vspace{-10pt}
\end{figure}

\textbf{Attack: Quantization substitution. }
We first compare outputs from $M_\text{spec}$ (full precision) and $M_\text{alt}$ (INT8 quantized) models. Using Llama-3/3.1-70B, with $L=50$ and $T=1$, and 25 multilingual Wikipedia prompts, we conduct 100 Monte Carlo simulations with 1000 permutations per test. Power is computed as the probability of rejecting $H_0{:} P_\text{actual}{=}P_\text{spec}$ when the entire output distribution came from $M_\text{alt}$.

\textbf{Attack: Randomized substitution. }
To simulate partial replacement, we construct mixtures of completions from $M_\text{spec}$ and $M_\text{alt}$ with substitution probability $p \in \{0.0,0.2,\dots,1.0\}$. We then repeat the same MMD test procedure to estimate how power changes as $p$ increases. This setup models a provider routing only a fraction of queries to the substituted model while serving the remainder from the specified one.

\textbf{Analysis. }
Figure~\ref{fig:Model_equality} shows that for smaller models (left panel), even full substitution by $M_\text{alt}$ remains statistically undetectable: power never rises beyond the significance level $\alpha$. For larger models (right panel), the test is more sensitive—power climbs with $p$ and reliably rejects $H_0$ when $p \geq 0.6$. However, for modest substitution ($p \leq 0.2$), rejection probability collapses to $\alpha$, making randomized substitution empirically indistinguishable from chance. These results imply that quantization substitution and partial randomized substitution are both viable evasion strategies, with the latter requiring only limited mixing to defeat auditing. Moreover, inference nondeterminism across providers can inflate MMD values even for identical models, risking false positives. As also noted by \citet{modelequalitytesting}, MMD-based auditing is only effective under strictly controlled local inference environments, limiting its practicality.

\subsubsection{Benchmark-based detection}

\textbf{Method. } 
An auditor may attempt benchmark-based detection (e.g., \citet{chen2023chatgptsbehaviorchangingtime}) by querying standardized benchmarks where baselines are well established. We evaluate four tasks: MMLU \citep{hendrycks2020measuring}, GSM8K \citep{cobbe2021training}, MATH \citep{hendrycks2021measuring}, and GPQA Diamond \citep{rein2024gpqa}. At decoding temperature \(T{=}0.5\), we repeat evaluation 10 times and report mean accuracy and standard deviation (Table~\ref{tab:benchmark_models_10runs}). MMLU is scored by restricting completions to multiple-choice tokens and averaging over 100 runs per query; free-form numeric tasks (GSM8K, MATH) are scored by regex matching or the \texttt{math-verify} library; GPQA is scored by multiple-choice accuracy.

\textbf{Attack 1: Quantization substitution. }
We compare quantized and full-precision variants across all four benchmarks. As shown in Table~\ref{tab:benchmark_models_10runs}, differences are small and typically within one standard deviation across repeated runs. This suggests that quantized models perform nearly indistinguishably from their full-precision counterparts, limiting the power of benchmark-only detection.

\begin{wraptable}{r}{0.62\textwidth}
    \centering
    \vspace{-8pt}
    \renewcommand{\arraystretch}{1.2}
    \resizebox{0.62\textwidth}{!}{
    \begin{tabular}{ccccc}
    \toprule
         \textbf{Model} & \textbf{MMLU} & \textbf{GSM8K} & \textbf{MATH} & \textbf{GPQA Diamond}\\ 
         \hline
         Meta-Llama-3-8B-Instruct & 62.69 $\pm$ 0.18 & 61.14 $\pm$ 3.47 & 20.65 $\pm$ 3.43 & 22.62 $\pm$ 0.22 \\
         Meta-Llama-3-8B-Instruct-FP8 & 62.43 $\pm$ 0.26 & 60.90 $\pm$ 4.10 & 14.91 $\pm$ 2.49 & 20.14 $\pm$ 0.24 \\
         \hline
         Meta-Llama-3-70B-Instruct & 78.05 $\pm$ 0.08 & 88.06 $\pm$ 1.44 & 35.69 $\pm$ 1.33 & 29.60 $\pm$ 0.51 \\
         Meta-Llama-3-70B-Instruct-FP8 & 77.88 $\pm$ 0.13 & 87.35 $\pm$ 1.24 & 35.75 $\pm$ 1.16 & 33.12 $\pm$ 0.30 \\
         \hline
         Gemma-2-9b-it & 71.86 $\pm$ 0.08 & 81.80 $\pm$ 1.35 & 33.41 $\pm$ 0.28 & 28.64 $\pm$ 2.97 \\
         Gemma-2-9b-it-FP8 & 71.92 $\pm$ 0.11 & 79.41 $\pm$ 1.14 & 32.53 $\pm$ 0.34 & 27.81 $\pm$ 3.22 \\
         \hline
         Qwen2-72B-Instruct & 82.18 $\pm$ 0.08 & 86.72 $\pm$ 1.00 & 37.39 $\pm$ 1.41 & 29.93 $\pm$ 2.71 \\
         Qwen2-72B-Instruct-FP8 & 81.98 $\pm$ 0.08 & 86.82 $\pm$ 0.97 & 37.67 $\pm$ 1.39 & 31.08 $\pm$ 1.96 \\
         \hline
         Mistral-7B-Instruct-v0.3 & 59.15 $\pm$ 0.10 & 35.90 $\pm$ 4.54 & 8.94 $\pm$ 1.24 & 21.60 $\pm$ 0.17 \\
         Mistral-7B-Instruct-v0.3-FP8 & 58.77 $\pm$ 0.13 & 32.20 $\pm$ 4.02 & 7.68 $\pm$ 1.23 & 22.72 $\pm$ 0.19 \\
         \bottomrule
    \end{tabular}
    }
    \caption{Mean $\pm$ standard deviation of benchmark results on different models at temperature \(T = 0.5\).}
    \label{tab:benchmark_models_10runs}
    \vspace{-10pt}
\end{wraptable}

\begin{figure}[h]
    \centering
    \vspace{-5pt}
    \includegraphics[width=0.7\linewidth]{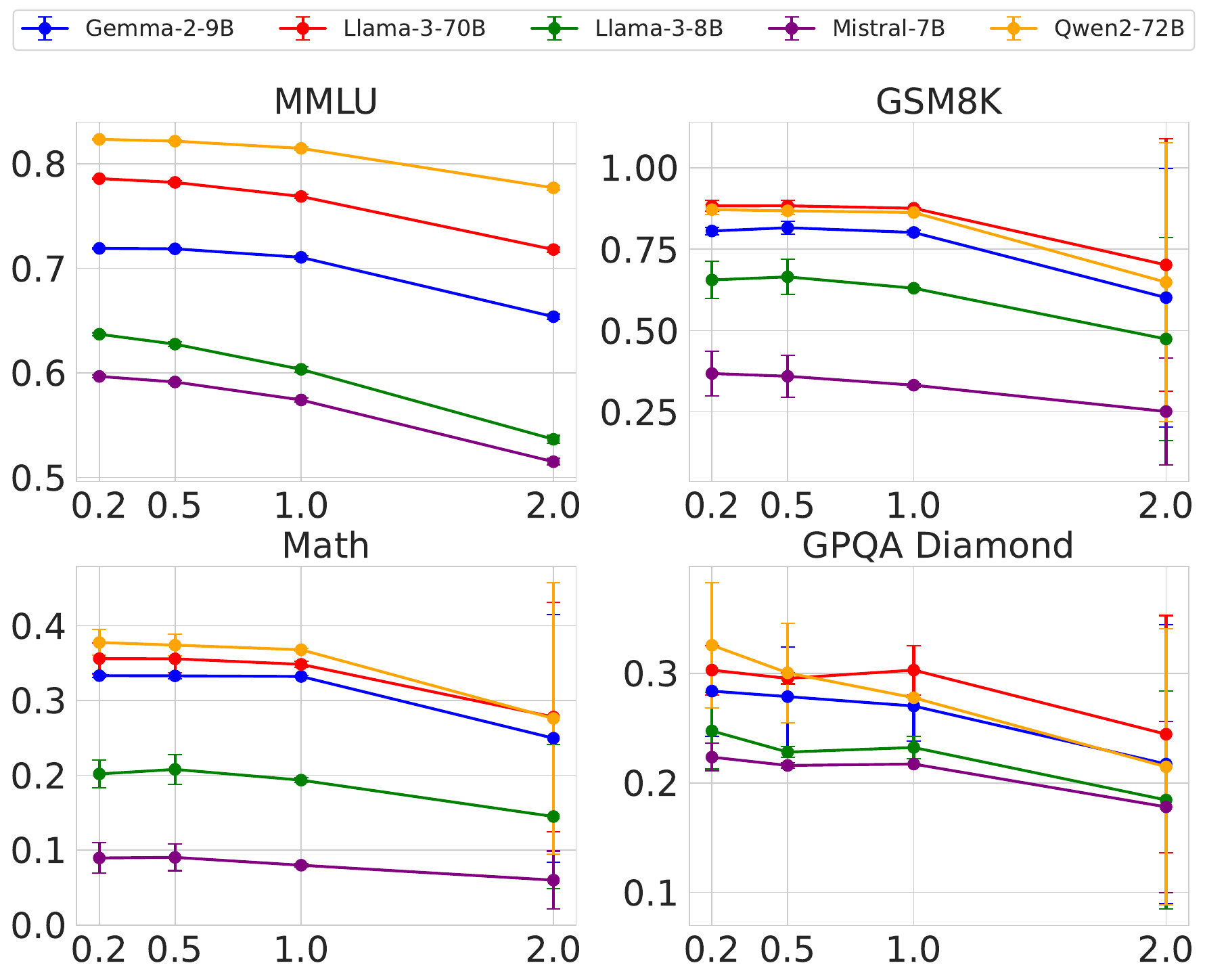}
    \vspace{-10pt}
    \caption{Benchmark accuracy versus decoding temperature across models and tasks.}
    \label{fig:benchmark}
\end{figure}

\textbf{Attack 2: Hiding temperature parameter.}
We also evaluate benchmark accuracy under varying decoding temperature (Figure~\ref{fig:benchmark}). For \(T \leq 1\), accuracy variance across runs is low, and single-run results are stable. However, if the provider does not disclose or allow control of temperature, observed accuracy drops could be misattributed—either to hidden parameter choices or to substitution—reducing audit reliability.

\textbf{Analysis. }
Figure~\ref{fig:benchmark} shows that benchmark accuracy degrades with higher $T$, and the size of this drop is comparable to quantization-induced differences. This means auditors testing $H_0{:} P_\text{actual}{=}P_\text{spec}$ cannot distinguish whether deviations from baseline arise from $M_\text{alt}$ substitution or from hidden parameter settings. Caching further complicates detection: if providers cache outputs for known benchmark queries, apparent accuracy may remain aligned with $M_\text{spec}$ even under substitution, suppressing deviations. Overall, benchmark-based detection can expose large-scale substitution, but smaller changes (e.g., quantization) are hidden by natural variability, and confounds like temperature or caching make $H_0$ rejection unreliable in practice.

\subsubsection{Greedy decoding outputs}

\textbf{Method. }
We test whether greedy decoding, which removes sampling randomness, yields reproducible completions across settings. Beyond substitution attacks, we also investigate how greedy decoding behaves in real-world auditing conditions, where inference is accessed only through APIs. This lets us assess not just sensitivity to quantization or fine-tuning, but also whether greedy decoding (and text-based verification more broadly) is robust when applied in deployment settings. Using 50 UltraChat queries, we compare local inference runs (Transformers 4.40.0 on H100 GPUs) against API providers’ outputs on the same Gemma-2-9B model. We also assess self-consistency across repeated API calls. Agreement is measured by exact match of the first \(k\) tokens, with \(k=20\) and \(k=100\).

\textbf{Attack: Quantization and fine-tuned substitution. }
For Llama-3-8B, we compare greedy outputs from the reference model against its quantized (FP8) and fine-tuned variants (e.g., models trained for other context length or on domain-specific corpora). While local inference across frameworks is largely consistent, both quantization and fine-tuning introduce significant token-level discrepancies, reducing agreement with the baseline sequence. This shows that greedy decoding can, in theory, detect within-family substitutions when the environment is controlled.

\begin{figure}[t]
    \centering
    \includegraphics[width=0.8\linewidth]{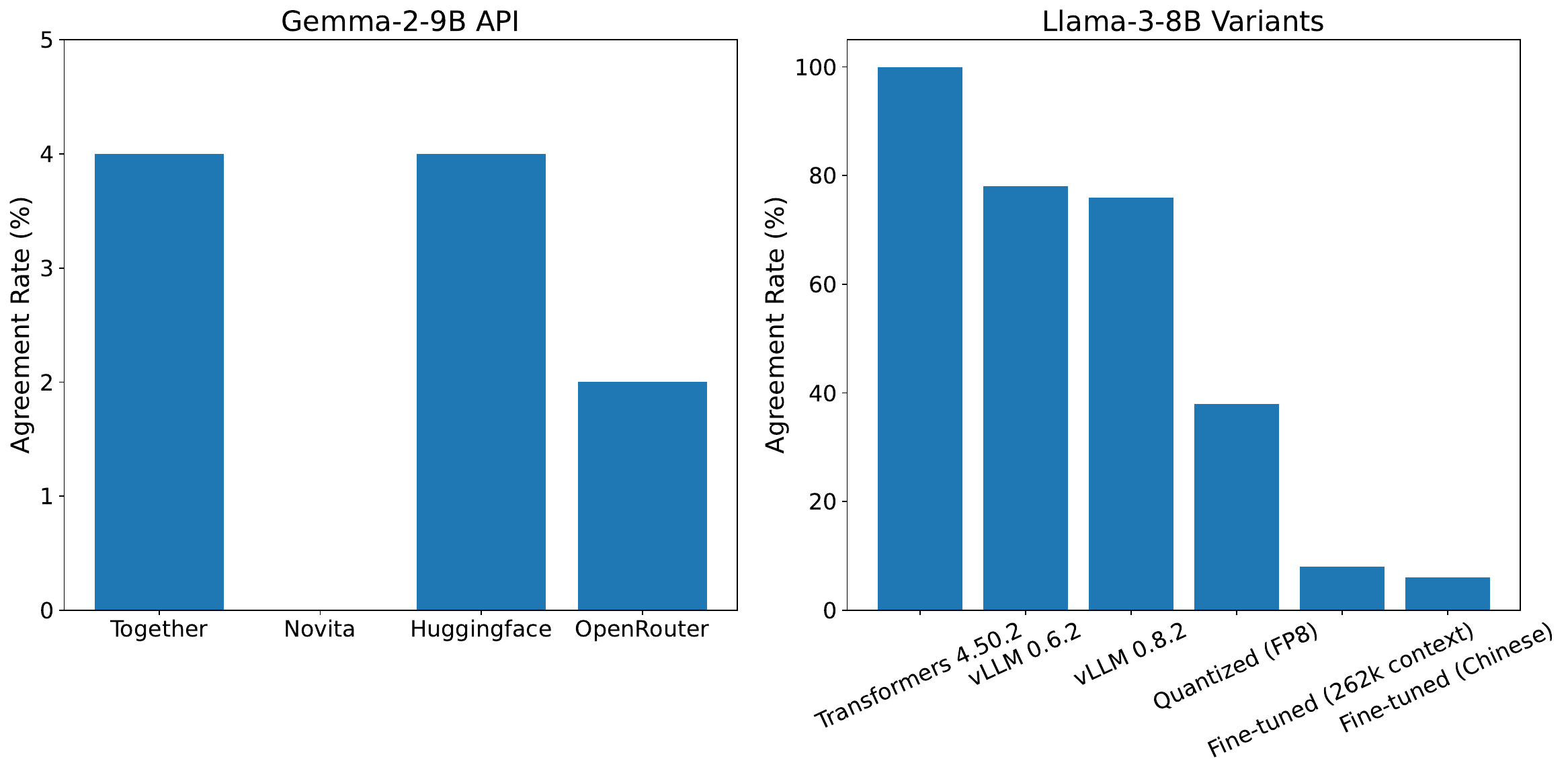}
    \vspace{-10pt}
    \caption{Greedy decoding agreement rate with local baseline (Transformers 4.40.0, H100). 
    Left: Gemma-2-9B APIs across providers. Right: Llama-3-8B variants including quantized and fine-tuned models.}
    \label{fig:greedy}
    \vspace{-10pt}
\end{figure}

\textbf{Analysis. }
Figure~\ref{fig:greedy} shows that greedy decoding does not ensure reproducibility: even with sampling disabled, agreement across provider APIs for Gemma-2-9B is below 5\%, and agreement for fine-tuned or quantized Llama-3-8B variants drops well below the baseline. Under $H_0: P_\text{actual}{=}P_\text{spec}$, greedy outputs should match exactly, but observed divergence is often larger than that induced by quantization or fine-tuning. This reflects inference nondeterminism \citep{he2025nondeterminism}, whose sources (e.g., backend variability in kernels, tokenization, or caching) are discussed in the next section. The key implication is that while greedy decoding can detect substitution in a controlled local environment, it fails to do so in realistic auditing settings where backend nondeterminism dominates, making $M_\text{spec}$ and $M_\text{alt}$ indistinguishable.

\subsubsection{Log probability verification}

\textbf{Method. }
In theory, logprobs comparison can be used to verify whether the served model matches a claimed reference: per-token log probabilities should align closely. In practice, the reliability of this approach depends on the magnitude of nondeterminism affecting the logprobs.

\begin{wrapfigure}{r}{0.55\textwidth}
    \centering
    \vspace{-10pt}
    \includegraphics[width=0.99\linewidth]{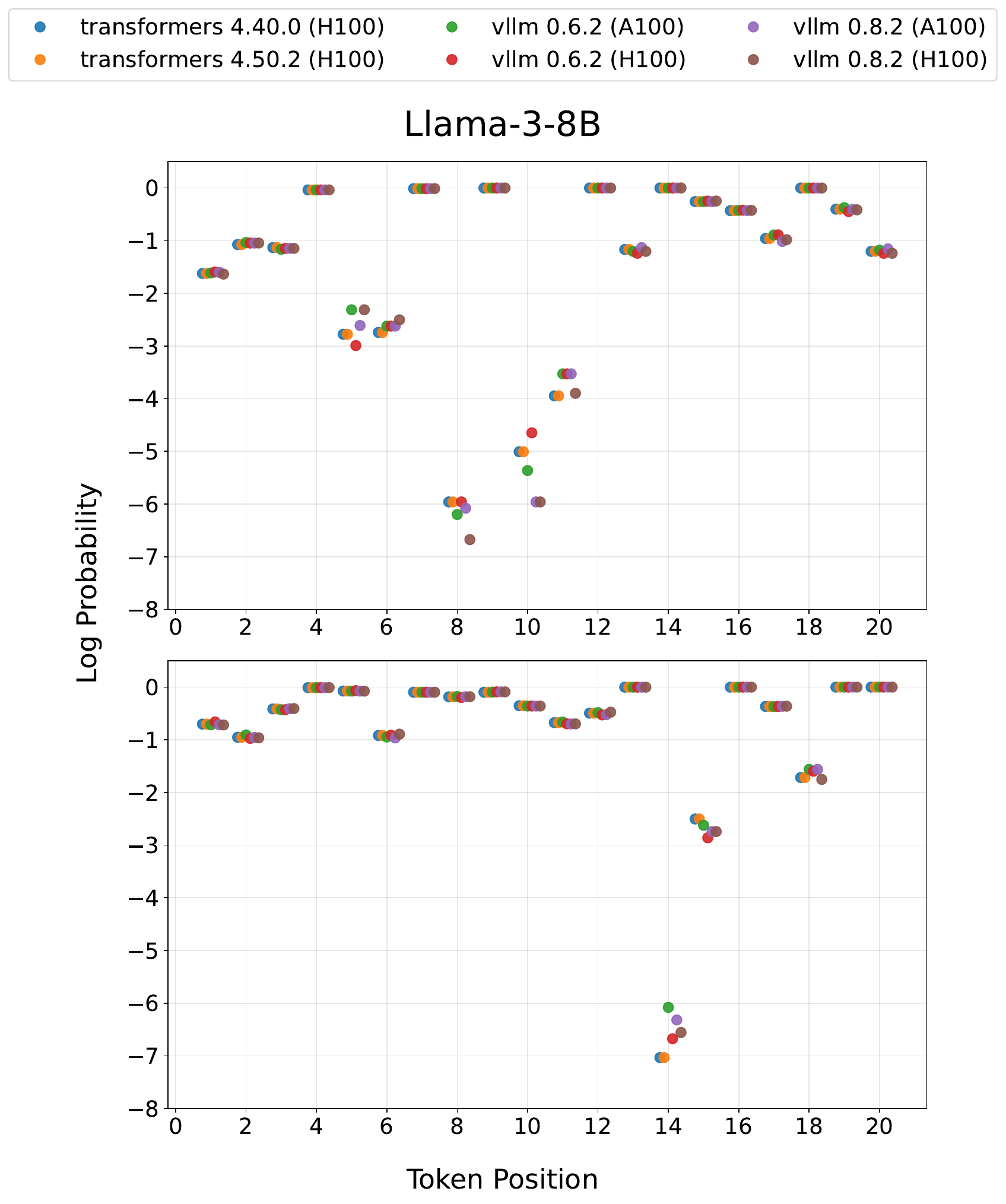}
    \vspace{-15pt}
    \caption{Log probabilities for the first 20 shared tokens under greedy decoding for UltraChat queries across frameworks/hardware.}
    \label{fig:logprobs}
    \vspace{-10pt}
\end{wrapfigure}

\textbf{Evaluation. }
To assess stability, we compare token-level log probabilities from greedy decoding on UltraChat queries, 
across multiple inference frameworks (vLLM \citep{kwon2023efficient} vs.\ Hugging Face Transformers \citep{wolf-etal-2020-transformers}), GPU types (H100, A100), 
and software versions. Figure~\ref{fig:logprobs} shows that even for the same model, logprob traces for the first 20 tokens can diverge across stacks. 
On the subset of completions where the generated tokens agreed, we further compare the magnitude of the assigned log probabilities and find minor variations across environments. 
Additional examples are given in Appendix~\ref{sec:embedding}.

\textbf{Analysis and weakness. }
Logprob traces often appear nearly identical across token positions in controlled settings, and the magnitude differences are still distinguishable when compared to quantized or fine-tuned variants. In production, however, batching and heterogeneous backends introduce much larger variance. Recent work \citep{he2025nondeterminism} shows that a key source of this instability is the lack of batch invariance in kernels such as RMSNorm, matrix multiplication, and attention: the same request may yield different results depending on how many other requests are batched concurrently or how prefill/chunking is scheduled. Even at $T=0$, greedy decoding can diverge after only tens of tokens, making logprob distributions very noisy. Prototype batch-invariant kernels exist but incur throughput costs, so current APIs typically remain nondeterministic. For auditors, this means logprob-based detection is vulnerable to false alarms unless requests are isolated or providers explicitly expose batch-invariant execution. In addition, this method requires a reference instance of $M_\text{spec}$ and API access to log probabilities. Stronger auditing methods that attempt to reconstruct embedding subspaces (Appendix~\ref{sec:embedding}) are generally infeasible under current API non-disclosure of full logprobs.

\subsection{Internal Activations Verification}
\label{sec:toploc}

\textbf{Method. }
TopLoc \citep{ong2025toploclocalitysensitivehashing} introduces a locality-sensitive hashing scheme over intermediate activations to produce compact proofs of correct execution. Here, locality-sensitive hashing (LSH) refers to mapping high-dimensional activation tensors into small fingerprints such that similar activations yield similar hashes, while even small unauthorized changes cause detectable differences. Instead of storing full tensors, TopLoc extracts the top-$k$ largest activation values and indices and encodes them as a polynomial congruence, yielding proofs that are robust to GPU nondeterminism and algebraic reorderings. This allows compact verification (about 100 bytes per tens of tokens) while maintaining accurate detection under adversarial substitution.

\textbf{Analysis and weakness. }
TopLoc can detect most model substitutions (e.g., quantized variants, fine-tuned models) even under different software and hardware settings, aligning with the behaviors observed in logprobs testing. However, verification still requires the auditor to recompute the same hidden activations, i.e., to run $M_{\text{spec}}$ locally or rely on an attested recomputation (e.g., a TEE), so it is infeasible without attestation for proprietary models. In addition, batching variance and speculative decoding could still remain problematic, as both can introduce enough instability in the forward pass or generated tokens to disrupt verification.

\subsection{Hardware-assisted verification: Trusted Execution Environment (TEE)}
\label{sec:tee}

\textbf{Method. } 
Trusted Execution Environments (TEEs), such as those provided by NVIDIA's Confidential Computing on Hopper and Blackwell GPUs \citep{nvidia}, offer a paradigm shift. A TEE creates a hardware-isolated enclave where both the model weights and the inference code are protected from the host system. The TEE can produce a cryptographic attestation report that includes measurements (hashes) of the loaded model and execution code. An end-user can cryptographically verify this report to gain absolute certainty that the intended model is running unmodified within the specified, trusted software environment.

\textbf{Evaluation. }
The primary concern with TEEs is performance overhead. We benchmark a vLLM inference endpoint for Llama-3-8B running on a single H100 GPU, both with and without a TEE enabled. We measure first-token latency and overall throughput under single-request and high-concurrency (64 requests) workloads.

\begin{wraptable}{r}{0.65\textwidth}
    \centering
    \vspace{-10pt}
    \renewcommand{\arraystretch}{1.2}
    \resizebox{0.99\linewidth}{!}{
    \begin{tabular}{ccccc}
    \toprule
Metric & \makecell{Concurrent\\requests \#} & TEE & no TEE & Overhead \\
\hline
\multirow{2}{*}{\makecell{First token latency\\(ms)}} & 1 & 71  & 65     & 9.23\%   \\
& 64 & 79 & 68 & 16.18\%  \\
\hline
\multirow{2}{*}{\makecell{Overall throughput\\(token/s)}} & 1 & 99  & 117    & 15.38\%  \\
& 64 & 943 & 971    & 2.88\%  \\
\bottomrule
    \end{tabular}
    }
    \vspace{-5pt}
    \caption{Performance comparison for a vLLM endpoint with/without TEE (Meta-Llama-3-8B-Instruct, single H100, TLS enabled).}
    \label{tab:seclm}
    \vspace{-10pt}
\end{wraptable}

\textbf{Analysis. }
Unlike software methods, TEEs are not vulnerable to computational attacks, as trust is anchored in hardware. The security guarantee is cryptographic, not statistical. Our evaluation in \Cref{tab:seclm} shows that this strong guarantee comes at a modest and practical cost. The overhead for first-token latency was 9-16\%, while the drop in overall throughput under a high-concurrency load was only 2.88\%. This demonstrates that TEEs offer a deployable, high-assurance solution to the model substitution problem, effectively closing the verification gap left by software-only techniques.

\section{Discussion}
\label{sec:discussion}

Our systematic evaluation reveals a clear hierarchy of verification capabilities:

\textbf{Text-only methods are insufficient. } 
Methods that only look at output distributions (classifiers, identity prompts, MMD, benchmarks) either need many queries, get confounded by benign variance, or can be evaded with some level of quantized model substitution. These approaches can give a rough signal but do not provide strong integrity guarantees. More generally, limited information access forces auditors to rely on large sample sizes to identify weak statistical patterns.

\textbf{Metadata-based methods are fragile. }
Logprobs offer more sensitivity: in principle, per-token likelihoods can reveal discrepancies with far fewer samples than text-only outputs. However, in practice, backend nondeterminism destroys this advantage: variance across frameworks, batching, and hardware is often larger than the effect of quantization or fine-tuning. TopLoc improves robustness by producing compact activation-based proofs that are less sensitive to such variability, but it has no guarantees under significantly different inference settings, and still requires local recomputation of $M_{\text{spec}}$, making it infeasible for proprietary deployments. In practice, both logprobs and TopLoc collapse without access to the reference model.

\textbf{Hardware-backed attestation is strongest. }
Trusted Execution Environments (TEEs) avoid these weaknesses: they guarantee integrity without exposing internals or requiring the auditor to hold the weights. Overheads are modest (\Cref{tab:seclm}), and attestation cryptographically binds the stack and model hash to the outputs. Adoption remains limited by operational complexity and weak provider incentives, but TEEs are currently the best compromise between verifiability, efficiency, and protection of proprietary models.

\textbf{Actionable recommendations: } (1) For users: Request attestation proofs when available; for critical applications, prefer providers offering TEE-backed services.
(2) For API providers: Consider TEE deployment for premium tiers; increase transparency through model versioning and metadata disclosure.
(3) For policymakers: Develop standards for API transparency and consider requiring attestation for regulated applications.

\section{Conclusion}
The model substitution problem represents a fundamental challenge in the current LLM ecosystem, where economic incentives misalign with user trust. Our work provides both a comprehensive analysis of why existing approaches fail and a concrete path forward through hardware attestation. As LLM APIs become increasingly critical infrastructure, ensuring computational integrity through TEEs may become as essential as HTTPS was for web security.

\section*{Ethics Statement}
This study focuses on auditing techniques for detecting model substitution in LLM APIs. 
All experiments are conducted using public datasets and benchmarks without involving human subjects or personal data, and API queries are issued within the provider's terms of service and rate limits. 
The attacks we discuss against text-output–based verification are not presented to encourage misuse, but to motivate the research community to develop stronger auditing methods under the same level of API access. 
Our goal is to improve transparency and trust in LLM services by highlighting weaknesses in current auditing approaches.

\bibliography{iclr2026_conference}

\begin{thebibliography}{41}
\providecommand{\natexlab}[1]{#1}
\providecommand{\url}[1]{\texttt{#1}}
\expandafter\ifx\csname urlstyle\endcsname\relax
  \providecommand{\doi}[1]{doi: #1}\else
  \providecommand{\doi}{doi: \begingroup \urlstyle{rm}\Url}\fi

\bibitem[{AMD}(2016)]{amd}
{AMD}.
\newblock {AMD Secure Encrypted Virtualization (SEV)}, 2016.
\newblock URL \url{https://www.amd.com/en/developer/sev.html}.
\newblock Accessed: 2025.

\bibitem[Anthropic(2023)]{anthropic_claude}
Anthropic.
\newblock Introducing claude, 2023.
\newblock URL \url{https://www.anthropic.com/news/introducing-claude}.

\bibitem[Bai et~al.(2023)Bai, Bai, Chu, Cui, Dang, Deng, Fan, Ge, Han, Huang, Hui, Ji, Li, Lin, Lin, Liu, Liu, Lu, Lu, Ma, Men, Ren, Ren, Tan, Tan, Tu, Wang, Wang, Wang, Wu, Xu, Xu, Yang, Yang, Yang, Yang, Yao, Yu, Yuan, Yuan, Zhang, Zhang, Zhang, Zhang, Zhou, Zhou, Zhou, and Zhu]{qwen}
Jinze Bai, Shuai Bai, Yunfei Chu, Zeyu Cui, Kai Dang, Xiaodong Deng, Yang Fan, Wenbin Ge, Yu~Han, Fei Huang, Binyuan Hui, Luo Ji, Mei Li, Junyang Lin, Runji Lin, Dayiheng Liu, Gao Liu, Chengqiang Lu, Keming Lu, Jianxin Ma, Rui Men, Xingzhang Ren, Xuancheng Ren, Chuanqi Tan, Sinan Tan, Jianhong Tu, Peng Wang, Shijie Wang, Wei Wang, Shengguang Wu, Benfeng Xu, Jin Xu, An~Yang, Hao Yang, Jian Yang, Shusheng Yang, Yang Yao, Bowen Yu, Hongyi Yuan, Zheng Yuan, Jianwei Zhang, Xingxuan Zhang, Yichang Zhang, Zhenru Zhang, Chang Zhou, Jingren Zhou, Xiaohuan Zhou, and Tianhang Zhu.
\newblock Qwen technical report.
\newblock \emph{arXiv preprint arXiv:2309.16609}, 2023.

\bibitem[Bao et~al.(2023)Bao, Zhao, Teng, Yang, and Zhang]{bao2023fast}
Guangsheng Bao, Yanbin Zhao, Zhiyang Teng, Linyi Yang, and Yue Zhang.
\newblock Fast-detectgpt: Efficient zero-shot detection of machine-generated text via conditional probability curvature.
\newblock In \emph{The Twelfth International Conference on Learning Representations}, 2023.

\bibitem[Carlini et~al.(2024)Carlini, Paleka, Dvijotham, Steinke, Hayase, Cooper, Lee, Jagielski, Nasr, Conmy, Wallace, Rolnick, and Tram\`{e}r]{stealingproductionlanguagemodel}
Nicholas Carlini, Daniel Paleka, Krishnamurthy~Dj Dvijotham, Thomas Steinke, Jonathan Hayase, A.~Feder Cooper, Katherine Lee, Matthew Jagielski, Milad Nasr, Arthur Conmy, Eric Wallace, David Rolnick, and Florian Tram\`{e}r.
\newblock Stealing part of a production language model.
\newblock In Ruslan Salakhutdinov, Zico Kolter, Katherine Heller, Adrian Weller, Nuria Oliver, Jonathan Scarlett, and Felix Berkenkamp (eds.), \emph{Proceedings of the 41st International Conference on Machine Learning}, volume 235 of \emph{Proceedings of Machine Learning Research}, 2024.

\bibitem[Chen et~al.(2023)Chen, Zaharia, and Zou]{chen2023chatgptsbehaviorchangingtime}
Lingjiao Chen, Matei Zaharia, and James Zou.
\newblock How is chatgpt's behavior changing over time?, 2023.
\newblock URL \url{https://arxiv.org/abs/2307.09009}.

\bibitem[Cobbe et~al.(2021)Cobbe, Kosaraju, Bavarian, Chen, Jun, Kaiser, Plappert, Tworek, Hilton, Nakano, et~al.]{cobbe2021training}
Karl Cobbe, Vineet Kosaraju, Mohammad Bavarian, Mark Chen, Heewoo Jun, Lukasz Kaiser, Matthias Plappert, Jerry Tworek, Jacob Hilton, Reiichiro Nakano, et~al.
\newblock Training verifiers to solve math word problems.
\newblock \emph{arXiv preprint arXiv:2110.14168}, 2021.

\bibitem[Ding et~al.(2023)Ding, Chen, Xu, Qin, Zheng, Hu, Liu, Sun, and Zhou]{ding2023enhancing}
Ning Ding, Yulin Chen, Bokai Xu, Yujia Qin, Zhi Zheng, Shengding Hu, Zhiyuan Liu, Maosong Sun, and Bowen Zhou.
\newblock Enhancing chat language models by scaling high-quality instructional conversations, 2023.
\newblock URL \url{https://arxiv.org/abs/2305.14233}.

\bibitem[Eyuboglu et~al.(2024)Eyuboglu, Goel, Desai, Chen, Monfort, R{\'e}, and Zou]{eyuboglu2024model}
Sabri Eyuboglu, Karan Goel, Arjun Desai, Lingjiao Chen, Mathew Monfort, Chris R{\'e}, and James Zou.
\newblock Model changelists: Characterizing updates to ml models.
\newblock In \emph{Proceedings of the 2024 ACM Conference on Fairness, Accountability, and Transparency}, pp.\  2432--2453, 2024.

\bibitem[Finlayson et~al.(2024)Finlayson, Ren, and Swayamdipta]{protectedllmsleak}
Matthew Finlayson, Xiang Ren, and Swabha Swayamdipta.
\newblock Logits of api-protected llms leak proprietary information, 2024.
\newblock URL \url{https://arxiv.org/abs/2403.09539}.

\bibitem[Gao et~al.(2024)Gao, Liang, and Guestrin]{modelequalitytesting}
Irena Gao, Percy Liang, and Carlos Guestrin.
\newblock Model equality testing: Which model is this api serving?, 2024.
\newblock URL \url{https://arxiv.org/abs/2410.20247}.

\bibitem[Ghosal et~al.(2023)Ghosal, Chakraborty, Geiping, Huang, Manocha, and Bedi]{ghosal2023towards}
Soumya~Suvra Ghosal, Souradip Chakraborty, Jonas Geiping, Furong Huang, Dinesh Manocha, and Amrit~Singh Bedi.
\newblock Towards possibilities \& impossibilities of ai-generated text detection: A survey.
\newblock \emph{arXiv preprint arXiv:2310.15264}, 2023.

\bibitem[Google(2023)]{google_gemini}
Google.
\newblock Introducing gemini: our largest and most capable ai model, 2023.
\newblock URL \url{https://blog.google/technology/ai/google-gemini-ai/}.

\bibitem[He \& Lab(2025)He and Lab]{he2025nondeterminism}
Horace He and Thinking~Machines Lab.
\newblock Defeating nondeterminism in llm inference.
\newblock \emph{Thinking Machines Lab: Connectionism}, 2025.
\newblock \doi{10.64434/tml.20250910}.
\newblock https://thinkingmachines.ai/blog/defeating-nondeterminism-in-llm-inference/.

\bibitem[Hendrycks et~al.(2020)Hendrycks, Burns, Basart, Zou, Mazeika, Song, and Steinhardt]{hendrycks2020measuring}
Dan Hendrycks, Collin Burns, Steven Basart, Andy Zou, Mantas Mazeika, Dawn Song, and Jacob Steinhardt.
\newblock Measuring massive multitask language understanding.
\newblock \emph{arXiv preprint arXiv:2009.03300}, 2020.

\bibitem[Hendrycks et~al.(2021)Hendrycks, Burns, Kadavath, Arora, Basart, Tang, Song, and Steinhardt]{hendrycks2021measuring}
Dan Hendrycks, Collin Burns, Saurav Kadavath, Akul Arora, Steven Basart, Eric Tang, Dawn Song, and Jacob Steinhardt.
\newblock Measuring mathematical problem solving with the math dataset.
\newblock \emph{arXiv preprint arXiv:2103.03874}, 2021.

\bibitem[Hong et~al.(2024)Hong, Duan, Zhang, Li, Xie, Lieberman, Diffenderfer, Bartoldson, Jaiswal, Xu, et~al.]{hong2024decoding}
Junyuan Hong, Jinhao Duan, Chenhui Zhang, Zhangheng Li, Chulin Xie, Kelsey Lieberman, James Diffenderfer, Brian Bartoldson, Ajay Jaiswal, Kaidi Xu, et~al.
\newblock Decoding compressed trust: Scrutinizing the trustworthiness of efficient llms under compression.
\newblock \emph{arXiv preprint arXiv:2403.15447}, 2024.

\bibitem[Hu et~al.(2023)Hu, Chen, and Ho]{hu2023radar}
Xiaomeng Hu, Pin-Yu Chen, and Tsung-Yi Ho.
\newblock Radar: Robust ai-text detection via adversarial learning.
\newblock \emph{Advances in neural information processing systems}, 36:\penalty0 15077--15095, 2023.

\bibitem[Huang et~al.(2025)Huang, Nasr, Angelopoulos, Carlini, Chiang, Choquette-Choo, Ippolito, Jagielski, Lee, Liu, Stoica, Tramer, and Zhang]{mitigatingadversarialmanipulation}
Yangsibo Huang, Milad Nasr, Anastasios Angelopoulos, Nicholas Carlini, Wei-Lin Chiang, Christopher~A. Choquette-Choo, Daphne Ippolito, Matthew Jagielski, Katherine Lee, Ken~Ziyu Liu, Ion Stoica, Florian Tramer, and Chiyuan Zhang.
\newblock Exploring and mitigating adversarial manipulation of voting-based leaderboards, 2025.
\newblock URL \url{https://arxiv.org/abs/2501.07493}.

\bibitem[{Intel}(2021)]{intel}
{Intel}.
\newblock {Intel Trust Domain Extensions (Intel TDX)}, 2021.
\newblock URL \url{https://www.intel.com/content/www/us/en/developer/tools/trust-domain-extensions/overview.html}.
\newblock Accessed: 2025.

\bibitem[Kirchenbauer et~al.(2023)Kirchenbauer, Geiping, Wen, Katz, Miers, and Goldstein]{kirchenbauer2023watermark}
John Kirchenbauer, Jonas Geiping, Yuxin Wen, Jonathan Katz, Ian Miers, and Tom Goldstein.
\newblock A watermark for large language models.
\newblock In \emph{International Conference on Machine Learning}, pp.\  17061--17084. PMLR, 2023.

\bibitem[Kwon et~al.(2023)Kwon, Li, Zhuang, Sheng, Zheng, Yu, Gonzalez, Zhang, and Stoica]{kwon2023efficient}
Woosuk Kwon, Zhuohan Li, Siyuan Zhuang, Ying Sheng, Lianmin Zheng, Cody~Hao Yu, Joseph Gonzalez, Hao Zhang, and Ion Stoica.
\newblock Efficient memory management for large language model serving with pagedattention.
\newblock In \emph{Proceedings of the 29th Symposium on Operating Systems Principles}, pp.\  611--626, 2023.

\bibitem[Mistral(2023)]{mistral2023mistral7b}
Mistral.
\newblock Mistral 7b: The best 7b model to date, apache 2.0, 2023.
\newblock URL \url{https://mistral.ai/news/announcing-mistral-7b/}.

\bibitem[Mitchell et~al.(2023)Mitchell, Lee, Khazatsky, Manning, and Finn]{mitchell2023detectgpt}
Eric Mitchell, Yoonho Lee, Alexander Khazatsky, Christopher~D Manning, and Chelsea Finn.
\newblock Detectgpt: Zero-shot machine-generated text detection using probability curvature.
\newblock In \emph{International Conference on Machine Learning}, pp.\  24950--24962. PMLR, 2023.

\bibitem[Murik \& Franke(2021)Murik and Franke]{measured-direct-boot}
Dov Murik and Hubertus Franke.
\newblock {Securing Linux VM boot with AMD SEV measurement}, 2021.
\newblock URL \url{https://static.sched.com/hosted_files/kvmforum2021/ed/securing-linux-vm-boot-with-amd-sev-measurement.pdf}.
\newblock Accessed: 2025.

\bibitem[{NVIDIA}(2023)]{nvidia}
{NVIDIA}.
\newblock {AI Security With Confidential Computing}.
\newblock NVIDIA Technical Blog, 2023.
\newblock URL \url{https://www.nvidia.com/en-us/data-center/solutions/confidential-computing/}.
\newblock Accessed: 2025.

\bibitem[Ong et~al.(2025)Ong, Di~Ferrante, Pazdera, Garner, Jaghouar, Basra, Hagemann, and Ryabinin]{ong2025toploclocalitysensitivehashing}
Jack~Min Ong, Matthew Di~Ferrante, Aaron Pazdera, Ryan Garner, Sami Jaghouar, Manveer Basra, Johannes Hagemann, and Max Ryabinin.
\newblock Toploc: A locality sensitive hashing scheme for trustless verifiable inference.
\newblock \emph{arXiv preprint arXiv:2501.16007}, 2025.
\newblock URL \url{https://arxiv.org/abs/2501.16007}.

\bibitem[OpenAI(2022)]{openai_chatgpt}
OpenAI.
\newblock Introducing chatgpt, 2022.
\newblock URL \url{https://openai.com/index/chatgpt/}.

\bibitem[Rein et~al.(2024)Rein, Hou, Stickland, Petty, Pang, Dirani, Michael, and Bowman]{rein2024gpqa}
David Rein, Betty~Li Hou, Asa~Cooper Stickland, Jackson Petty, Richard~Yuanzhe Pang, Julien Dirani, Julian Michael, and Samuel~R Bowman.
\newblock Gpqa: A graduate-level google-proof q\&a benchmark.
\newblock In \emph{First Conference on Language Modeling}, 2024.

\bibitem[Sam et~al.(2025)Sam, Finzi, and Kolter]{sam2025predicting}
Dylan Sam, Marc Finzi, and J~Zico Kolter.
\newblock Predicting the performance of black-box llms through self-queries.
\newblock \emph{arXiv preprint arXiv:2501.01558}, 2025.

\bibitem[Sun et~al.(2024{\natexlab{a}})Sun, Li, and Zhang]{sun2024zkllm}
Haochen Sun, Jason Li, and Hongyang Zhang.
\newblock zkllm: Zero knowledge proofs for large language models.
\newblock In \emph{Proceedings of the 2024 on ACM SIGSAC Conference on Computer and Communications Security}, pp.\  4405--4419, 2024{\natexlab{a}}.

\bibitem[Sun et~al.(2025)Sun, Yin, Xu, Kolter, and Liu]{idiosyncrasieslargelanguagemodels}
Mingjie Sun, Yida Yin, Zhiqiu Xu, J.~Zico Kolter, and Zhuang Liu.
\newblock Idiosyncrasies in large language models.
\newblock \emph{arXiv preprint arXiv:2502.12150}, 2025.

\bibitem[Sun et~al.(2024{\natexlab{b}})Sun, Li, Zhang, Jin, and Zhang]{sun2024svip}
Yifan Sun, Yuhang Li, Yue Zhang, Yuchen Jin, and Huan Zhang.
\newblock Svip: Towards verifiable inference of open-source large language models.
\newblock \emph{arXiv preprint arXiv:2410.22307}, 2024{\natexlab{b}}.

\bibitem[TogetherAI(2023)]{together_inference}
TogetherAI.
\newblock Announcing together inference engine – the fastest inference available, 2023.
\newblock URL \url{https://www.together.ai/blog/together-inference-engine-v1}.

\bibitem[Touvron et~al.(2023)Touvron, Lavril, Izacard, Martinet, Lachaux, Lacroix, Rozi{\`e}re, Goyal, Hambro, Azhar, et~al.]{touvron2023llama}
Hugo Touvron, Thibaut Lavril, Gautier Izacard, Xavier Martinet, Marie-Anne Lachaux, Timoth{\'e}e Lacroix, Baptiste Rozi{\`e}re, Naman Goyal, Eric Hambro, Faisal Azhar, et~al.
\newblock Llama: Open and efficient foundation language models.
\newblock \emph{arXiv preprint arXiv:2302.13971}, 2023.

\bibitem[Wolf et~al.(2020)Wolf, Debut, Sanh, Chaumond, Delangue, Moi, Cistac, Rault, Louf, Funtowicz, Davison, Shleifer, von Platen, Ma, Jernite, Plu, Xu, Scao, Gugger, Drame, Lhoest, and Rush]{wolf-etal-2020-transformers}
Thomas Wolf, Lysandre Debut, Victor Sanh, Julien Chaumond, Clement Delangue, Anthony Moi, Pierric Cistac, Tim Rault, Rémi Louf, Morgan Funtowicz, Joe Davison, Sam Shleifer, Patrick von Platen, Clara Ma, Yacine Jernite, Julien Plu, Canwen Xu, Teven~Le Scao, Sylvain Gugger, Mariama Drame, Quentin Lhoest, and Alexander~M. Rush.
\newblock Transformers: State-of-the-art natural language processing.
\newblock In \emph{Proceedings of the 2020 Conference on Empirical Methods in Natural Language Processing: System Demonstrations}, pp.\  38--45, Online, October 2020. Association for Computational Linguistics.
\newblock URL \url{https://www.aclweb.org/anthology/2020.emnlp-demos.6}.

\bibitem[Xie et~al.(2025)Xie, Lu, Fang, Wang, Zhang, Jia, Song, and Zhang]{xie2025zkpytorch}
Tiancheng Xie, Tao Lu, Zhiyong Fang, Siqi Wang, Zhenfei Zhang, Yongzheng Jia, Dawn Song, and Jiaheng Zhang.
\newblock zkpytorch: A hierarchical optimized compiler for zero-knowledge machine learning.
\newblock \emph{Cryptology ePrint Archive}, 2025.

\bibitem[Yang et~al.(2023)Yang, Pan, Zhao, Chen, Petzold, Wang, and Cheng]{yang2023survey}
Xianjun Yang, Liangming Pan, Xuandong Zhao, Haifeng Chen, Linda Petzold, William~Yang Wang, and Wei Cheng.
\newblock A survey on detection of llms-generated content.
\newblock \emph{arXiv preprint arXiv:2310.15654}, 2023.

\bibitem[Yang et~al.(2024)Yang, Cheng, Wu, Petzold, Wang, and Chen]{yang2024dna}
Xianjun Yang, Wei Cheng, Yue Wu, Linda~Ruth Petzold, William~Yang Wang, and Haifeng Chen.
\newblock {DNA-GPT:} divergent n-gram analysis for training-free detection of gpt-generated text.
\newblock In \emph{The Twelfth International Conference on Learning Representations, {ICLR} 2024, Vienna, Austria, May 7-11, 2024}. OpenReview.net, 2024.
\newblock URL \url{https://openreview.net/forum?id=Xlayxj2fWp}.

\bibitem[Zhao et~al.(2024{\natexlab{a}})Zhao, Ananth, Li, and Wang]{zhao2024provable}
Xuandong Zhao, Prabhanjan~Vijendra Ananth, Lei Li, and Yu{-}Xiang Wang.
\newblock Provable robust watermarking for ai-generated text.
\newblock In \emph{The Twelfth International Conference on Learning Representations, {ICLR} 2024, Vienna, Austria, May 7-11, 2024}. OpenReview.net, 2024{\natexlab{a}}.
\newblock URL \url{https://openreview.net/forum?id=SsmT8aO45L}.

\bibitem[Zhao et~al.(2024{\natexlab{b}})Zhao, Gunn, Christ, Fairoze, Fabrega, Carlini, Garg, Hong, Nasr, Tram{\`e}r, Jha, Li, Wang, and Song]{zhao2024sok}
Xuandong Zhao, Sam Gunn, Miranda Christ, Jaiden Fairoze, Andres Fabrega, Nicholas Carlini, Sanjam Garg, Sanghyun Hong, Milad Nasr, Florian Tram{\`e}r, Somesh Jha, Lei Li, Yu-Xiang Wang, and Dawn Song.
\newblock Sok: Watermarking for ai-generated content.
\newblock \emph{arXiv preprint arXiv:2411.18479}, 2024{\natexlab{b}}.

\end{thebibliography}
\bibliographystyle{iclr2026_conference}

\appendix
\section*{Use of Large Language Models (LLMs)}
We acknowledge the use of LLMs as general-purpose assist tools during the preparation of this manuscript. Specifically, an LLM was utilized for paraphrasing certain sentences and for correcting grammatical errors and improving sentence structure throughout the text. This assistance was primarily aimed at enhancing the clarity and readability of our writing. The authors take full responsibility for the content of this paper, including any generated text, and confirm that all scientific ideas, experimental design, results, and conclusions are original contributions of the human authors. LLMs were not considered as authors or contributors to the research ideation.

\section{Additional details}

\begin{table}[h]
    \centering
    \setlength{\tabcolsep}{8pt}
    \renewcommand{\arraystretch}{1.3}
    \resizebox{\textwidth}{!}{
    \begin{tabular}{llll}
    \toprule
        \textbf{Service Provider} & \textbf{Reference documentation} \\
        \midrule
        Anyscale & \url{https://docs.anyscale.com/endpoints/text-generation/logprobs/} \\
        Together.ai & \url{https://docs.together.ai/docs/logprobs}  \\
        Hugging Face & \url{https://huggingface.co/docs/api-inference/tasks/chat-completion\#request}  \\
        AWS Bedrock & \url{https://docs.aws.amazon.com/bedrock/latest/userguide/model-parameters.html} \\
        Nebius AI & \url{https://docs.nebius.com/studio/inference}  \\
        Vertex AI & \url{https://cloud.google.com/vertex-ai/generative-ai/docs/multimodal/content-generation-parameters} \\
        Mistral & \url{https://docs.mistral.ai/api/\#operation/createChatCompletion} \\
        DeepSeek & \url{https://api-docs.deepseek.com/api/create-completion} \\
        OpenAI & \url{https://platform.openai.com/docs/api-reference/chat/create\#chat-create-top_logprobs} \\
        Cohere & \url{https://docs.cohere.com/v2/reference/chat\#request.body.logprobs} \\
        Anthropic & \url{https://docs.anthropic.com/en/api/messages} \\
        \bottomrule
    \end{tabular}
    }
    \caption{LLM API service providers documentations.}
    \label{tab:service_providers_doc}
\end{table}

\begin{figure}[h]
    \centering
    \includegraphics[width=0.85\linewidth]{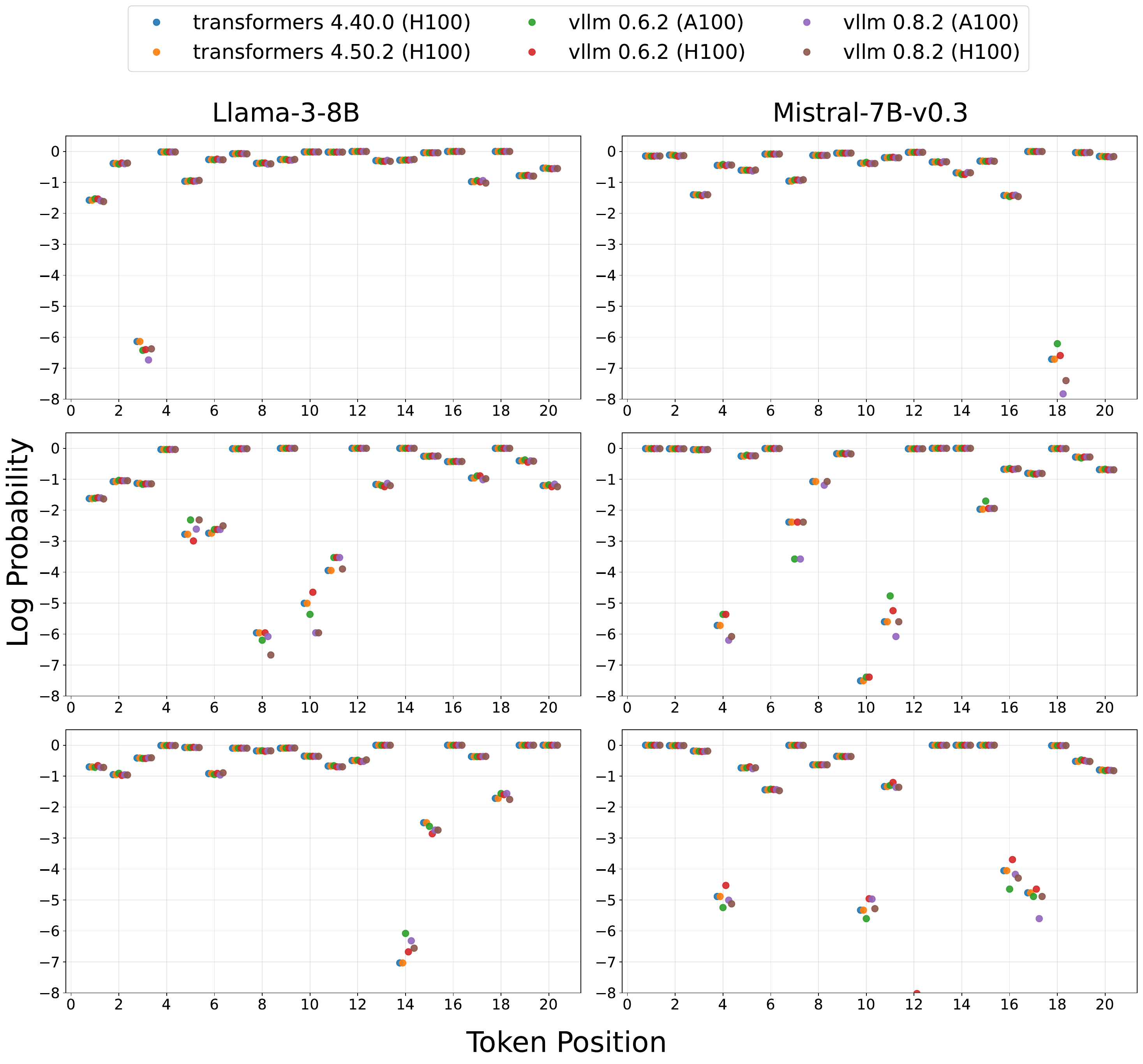}
    \caption{Log probability of generating first 20 shared tokens under greedy decoding for UltraChat Queries under different environment.}
    \label{fig:Logprobs2}
\end{figure}

\section{Model stealing and embedding verification}
\label{sec:embedding}

\textbf{Background. }
Researchers have explored extracting model information (e.g., weights, architecture details) from black-box APIs \citep{stealingproductionlanguagemodel, protectedllmsleak}. While successful extraction could reveal substitutions, these methods often require a vast number of queries, significant computational resources, and may still yield incomplete information, making them impractical for routine auditing by typical users. \citet{protectedllmsleak} specifically demonstrate how logits analysis could expose model details, which we consider as a potential verification method for detecting model substitutions.

\textbf{Method. }
The embedding size detection relies on the fact that logits generated by an LLM are restricted to a \(d\)-dimensional subspace of the full \(v\)-dimensional output space where \(d\) is the hidden embedding size. So the auditor can analyze a set of logit outputs from different prompts and reconstruct this subspace and the hidden embedding size.

Given a set of \(n\) output logit vectors \(\{\ell_i\}_{i=1}^{n}\) (or log probabilities, since it preserves linear relationship up to a constant addition), we construct a logit matrix \(L\) with each vector as its columns. Applying SVD to \(L\) yields \(L = U \Sigma V^T,\) where \(U \in \mathbb{R}^{v \times v}\), \(\Sigma \in \mathbb{R}^{v \times n}\) (with singular values \(\lambda_1 \geq \lambda_2 \geq \dots\)), and \(V \in \mathbb{R}^{n \times n}\). The magnitude of the singular values will drastically decreases after the first \(d\) dimensions. Formally, the embedding size \(d\) can be identified by detecting the singular value index at which the magnitude drops the most:
$
d = \arg\max_i (\log \lambda_i - \log \lambda_{i+1}),
$
Once identified, the \(d\)-dimensional subspace is constructed using the first \(d\) left singular vectors from \(U\):
$
U_d = [u_1, u_2, \dots, u_d] \in \mathbb{R}^{v \times d},
$ forming a unique signature of the model. Empirical evaluations confirm that this method accurately recovers the embedding dimensions. Furthermore, monitoring this subspace allows auditors to detect subtle changes, such as hidden system prompts, fine-tuning, or entire model substitutions from the service provider.

\textbf{Weakness. }
This verification method is impractical because no well-known service provider gives full log probability access.

\section{Build an LLM inference API endpoint with TEE}
\label{sec:appendix-tee}
In this section, we provide a solution for building an LLM inference API endpoint with TEE.
Considering the main program is still running on the CPU, the GPU TEE technology needs to be paired with the VM-based CPU TEE (e.g., Intel TDX \citep{intel}, AMD SEV-SNP \citep{amd}) to work properly. The latest CPU TEE can launch a confidential virtual machine inside and ensure the integrity and confidentiality of the VM's memory, preventing the host which is controlled by the API provider from manipulating the data inside the confidential VM. The GPU TEE (e.g., NVIDIA Confidential Computing \citep{nvidia}) can build a secure communication channel between the CPU and GPU and ensure the integrity and confidentiality of the data on the GPU memory, thus we can add the GPU to the trusted computing base (TCB). By combining these two TEEs, we can ensure the integrity and confidentiality of programs and data processed inside them.

To ensure integrity and confidentiality, the program inside the TEE needs to generate a key pair for identifying itself and establishing secure communication channels with end users. This key can help end users verify that they are connecting to the program inside the TEE and ensure the API providers cannot interfere the communication traffic. Once the data is transferred to TEE, the data processing inside TEE will be encrypted by hardware to ensure integrity and confidentiality.

One last question is how we can provide proof of the integrity of the inference program and add this proof to the attestation report. First, the VM-based CPU TEE ensures memory protection and the measured direct boot \citep{measured-direct-boot} extends the measurement to kernel, initrd, and kernel command line. At this point, the measurement value in the attestation report can validate the integrity of the kernel, initrd, and kernel cmdline. However, ensuring the integrity of the operating system and program on the disk requires additional steps. To achieve this, we include a program into the initrd that verifies and encrypts disk data. This action incorporates the disk into the trusted computing base (TCB).

In this setting, the LLM inference program needs to be open-sourced for verification, while the model can remain private if needed. The inference program inside the TEE can verify the hash of the model weights and ensure that every time a model with a specific name is used, it has the same hash. For open-sourced models, it is easy for anyone to calculate the desired hash. For proprietary models, the name serves as an alias for specific model weights, if we can confirm that it always has the same hash value, we can consider it to be integrous.

\begin{figure}
    \centering
    \includegraphics[width=\linewidth]{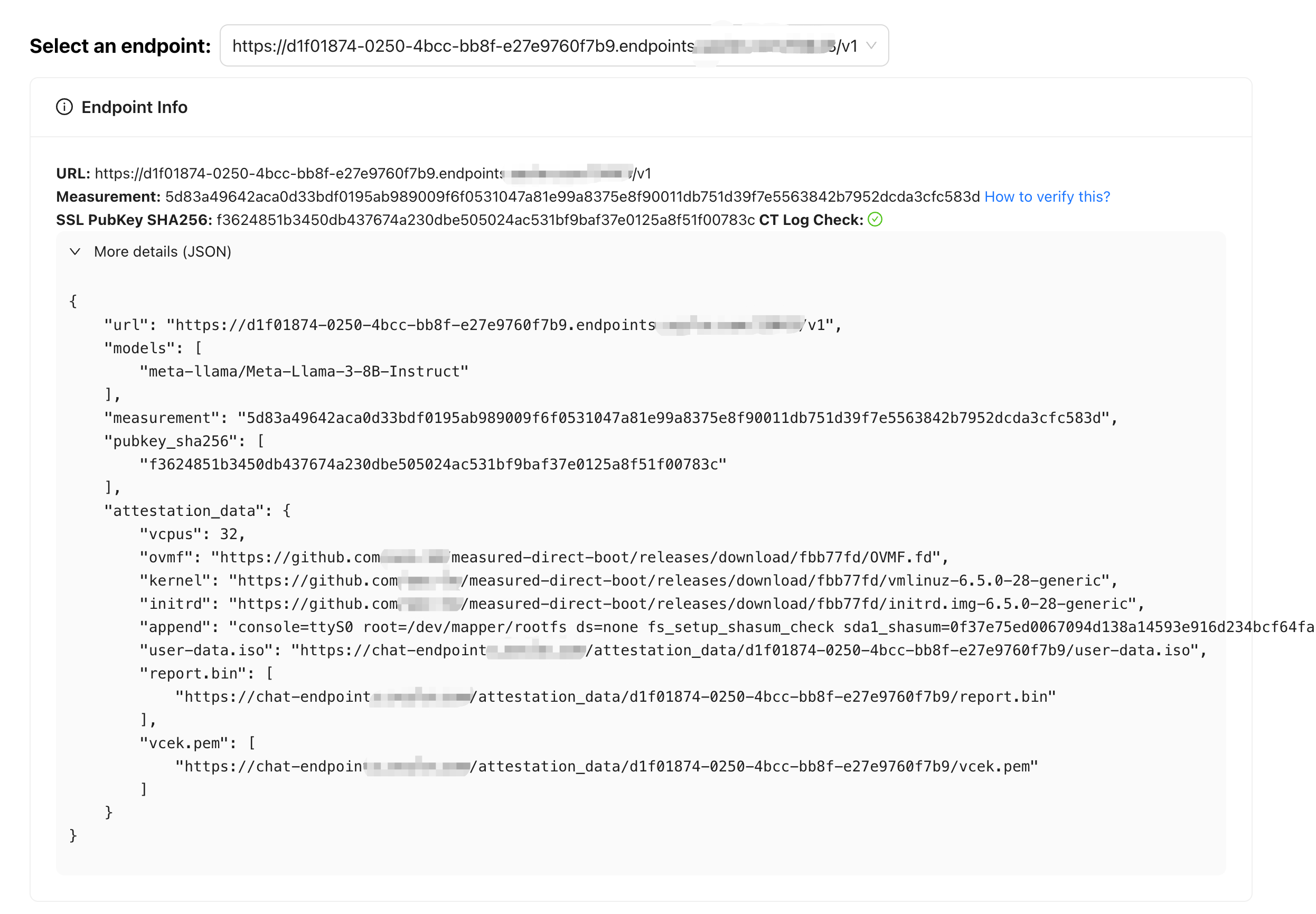}
    \caption{A screenshot of the API inference endpoint with TEE.}
    \label{fig:seclm}
\end{figure}

As we can see in Figure \ref{fig:seclm}, the program is transparent and verifiable to the end users. End users can check the attestation report and verify the signature on it to ensure that the API provider is running the desired program.

\end{document}